\documentclass[final]{cvpr}

\usepackage{mathptmx}
\usepackage{graphicx}
\usepackage{graphicx}
\usepackage{amsmath}
\usepackage{amssymb}
\usepackage{subfig}
\usepackage{caption}
\usepackage{booktabs}
\usepackage{multirow}
\usepackage{bm}
\makeatletter
\@namedef{ver@everyshi.sty}{}
\makeatother
\usepackage{tikz}
\usepackage{appendix}

\usepackage[pagebackref=true,breaklinks=true,colorlinks,bookmarks=false]{hyperref}

\def\cvprPaperID{****} 
\def\confYear{CVPR 2021}

\begin{document}

\title{Brittle Features May Help Anomaly Detection}

\author{Kimberly T. Mai, Toby Davies, Lewis D. Griffin\\
University College London\\
London, UK\\
{\tt\small \{kimberly.mai, toby.davies, l.griffin\}@ucl.ac.uk}
}

\maketitle

\begin{abstract}
One-class anomaly detection is challenging. A representation that clearly distinguishes anomalies from normal data is ideal, but arriving at this representation is difficult since only normal data is available at training time. We examine the performance of representations, transferred from auxiliary tasks, for anomaly detection. Our results suggest that the choice of representation is more important than the anomaly detector used with these representations, although knowledge distillation can work better than using the representations directly. In addition, separability between anomalies and normal data is important but not the sole factor for a good representation, as anomaly detection performance is also correlated with more adversarially brittle features in the representation space. Finally, we show our configuration can detect 96.4\% of anomalies in a genuine X-ray security dataset, outperforming previous results. 
\end{abstract}

\section{Introduction}


Anomaly detection is the task of identifying instances that deviate from typical patterns in data. Effective anomaly detection is critical for many applications, for example:  identifying suspicious items in X-ray baggage scans \cite{griffin1}, detecting biomarkers in medical imaging \cite{anogan}, or flagging potential defects in products \cite{mvtec}. It is evident that the definition of an anomaly can vary drastically between applications. Consequently, domain knowledge must be used to inform how the anomaly detector is built. 

The straightforward approach would be to train a supervised classifier to distinguish between normal and anomalous training examples. However, in addition to the scarcity of anomalous samples suitable for training, there is no guarantee that the anomalies used for training encompass all possible types of anomalies that could manifest. Adversaries may look to take advantage of this shortcoming. For example, an adversary looking to smuggle illicit goods may learn about certain arrangements in baggage which are commonly flagged by detection systems and may develop more sophisticated techniques to conceal their goods so that the contents of their baggage appears benign \cite{andrews1}. 

One-class methods are an alternative approach. In this scenario, only normal data is available at training time. The model learns representations of the normal data and does not make any a priori assumptions about anomalies. At test time, the model is presented with both normal and anomalous data, and anomaly detection can be performed by using density estimation to evaluate how likely a test datum originates from the normal data distribution \cite{bishop}.

In practice however, generative models trained for one-class anomaly detection tend to assign higher anomaly scores to normal data compared to anomalies \cite{Nalisnick_Matsukawa_Teh_Gorur_Lakshminarayanan_2019}. This has been attributed to both poor calibration between the true underlying distribution and the fitted distribution, and the curse of dimensionality \cite{ren,schirrmeister, wang}. 

In contrast, it has been shown that discriminative self-supervised methods are capable of learning good representations that perform well in downstream tasks such as classification \cite{simclr, rotnet, BYOL, pathak, zhang}. Therefore, we pivot to the use of these discriminative representations for anomaly detection on images primarily through the means of knowledge distillation.

As a substitute for human-annotated labels, a student network (student) which only sees normal training data is trained to match the internal representations of a frozen teacher network (teacher). Leveraging the idea that regression models extrapolate poorly to unseen data, we expect the representations of the student and teacher to differ more on anomalous images as opposed to normal images \cite{burda}. This should be indicated through larger regression errors for anomalies, so mean squared error (MSE) can be used as the anomaly score. Methods that have previously built on the idea of failure-to-extrapolate have fixed the teacher representations without analysing which auxiliary tasks, that are used to derive such representations, are better suited for anomaly detection  \cite{uninformed, burda, Ciosek2020Conservative}. 

In this paper, we study the effect of different teacher representations on anomaly detection performance. These representations are derived from a variety of sources, including: random representations, representations transferred from different classification tasks, discriminative self-supervised tasks, and autoencoders. These representations were evaluated on different image datasets, including an X-ray security dataset containing staged threats \cite{griffin1} for which we outperform the previous best anomaly detection scores. 

Our results suggest that representations vary in their effectiveness for anomaly detection, and those with `brittle' features, which are shown to be susceptible to adversarial perturbations, may perform better. To summarise, our findings are as follows:

\begin{itemize}
    \item Representations are more important than the anomaly detection method.
    \item In the instances where the representation is suitable for anomaly detection, knowledge distillation outperforms anomaly detectors that use the representation directly.
    \item Good representations not only require clear separation of anomalies and normal data in this space but also the use of brittle features.
\end{itemize}
\section{Related Work}

\textbf{Self-supervised models} for anomaly detection use an auxiliary task to learn a representation of the normal data.

Reconstruction-type methods such as PCA, autoencoders, and GANs \cite{ganomaly, sakurada, anogan} learn a set of bases for the normal data. As it is expected these bases cannot sufficiently represent anomalies, reconstruction error is used as the anomaly score. This is not always observed in practice. For example, if the anomalies contain simpler features compared to the normal data, the bases may be sufficient to reconstruct the anomalies and the resultant reconstruction errors may be smaller.

Other auxiliary tasks involve contrastive learning or predicting transformations applied to input data.  It is expected that the distance of anomalous representations to typical representations in this space is greater compared to normal test representations \cite{geom, tack2020csi}. However, such techniques use transformations on the input data to learn the representation space, hence knowledge of appropriate transformations that do not erode the features that distinguish normal and anomalous data is required.

\textbf{Transfer learning} from pre-trained models can also be leveraged for anomaly detection. Previous approaches typically feed the representations through a shallow model \cite{andrews3, lee_mahalanobis} or apply feature engineering such as binarisation to enhance the differences between normal and anomalous features \cite{griffin1}.

\textbf{Knowledge distillation} has been used for anomaly segmentation in images \cite{uninformed} and to aid exploration in reinforcement learning \cite{burda}. However, the representations used have been selected without analysing which ones work better. Popular teacher representations for anomaly detection on images are randomly initialised networks or networks pre-trained to classify ImageNet.

Knowledge distillation has also been used to augment the performance of supervised classifiers. One explanation for improved performance is that the logits provide more information about how the classes are related compared to labels \cite{kd, journals/corr/RomeroBKCGB14}.

\textbf{Adversarial Robustness}: Neural networks tend to be susceptible to adversarial perturbations. As neural networks are typically optimised using first-order methods, it has been shown that large gradient norms are indicative of increased adversarial vulnerability in a network \cite{pmlr-v97-simon-gabriel19a}. The occurrence of adversarial examples suggests neural networks rely on brittle features for learning representations \cite{journals/corr/GoodfellowSS14}, and it has been argued that these brittle features are vital for achieving good generalisation performance \cite{ilyas, conf/iclr/TsiprasSETM19}.
\section{How Representations were Compared}

\subsection{Datasets}
We evaluate anomaly detection performance on Cats vs. Dogs (CvD) \cite{asirra-a-captcha-that-exploits-interest-aligned-manual-image-categorization}, CIFAR-10 (CIFAR), the Plant Pathology 2020 dataset (Plant Path.) \cite{thapa2020plant}, and the X-ray parcel dataset (X-ray) used in Griffin \etal \cite{griffin1}.

\textbf{Cats vs. Dogs}: Cats vs. Dogs contains 25,000 images of cats and dogs, split equally between the two classes. For each class, we allocated 10,000 images into the training set and the remaining 2,500 images were allocated into the test set. Each image was resized to $32 \times 32$. The anomaly detectors were trained on one class during the training stage and evaluated on both classes in the testing stage. 

\textbf{CIFAR-10}: We examine both tight and wide normal classes as per Kim \etal \cite{rapp}: 
\begin{itemize}
    \item \textbf{Unimodal}: One class is chosen as the normal class and the remaining nine are designated as anomalous.
    \item \textbf{Multimodal}: One class is chosen as the anomalous class and the remaining nine are designated as normal.
\end{itemize}
This setup is repeated for all 10 classes, hence there are 20 configurations in total.

\textbf{Plant Pathology}: The Plant Pathology 2020 dataset  contains 3,651 RGB images of apple leaves taken under different light, angle, and noise conditions. Of those, a total of 1,821 images were labelled and categorised into one of four categories: healthy (516), containing apple scabs (592), containing cedar apple rust (622), or containing multiple diseases (91). 80\% of the healthy images were used for training the anomaly detector and the remaining 20\% were used for testing. We sampled an equal number of images from the other classes to serve as anomalies at the testing stage. Each image was resized to $224 \times 224$.

\begin{figure}[h]
    \centering
\subfloat{\label{xfig:a}\includegraphics[width=.22\textwidth, height=1.5cm]{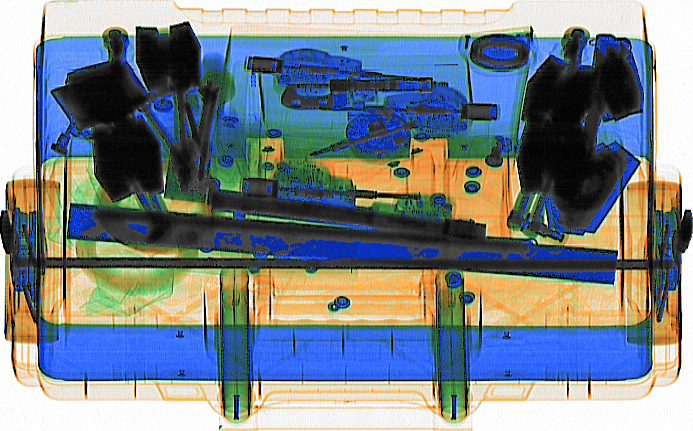}}\hfill
\subfloat{\label{xfig:b}\includegraphics[width=.22\textwidth, height=1.5cm]{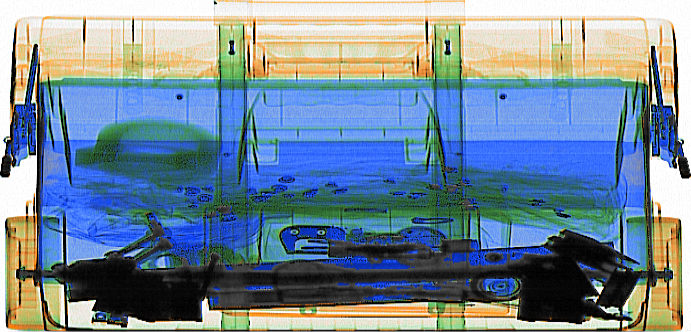}}\hfill
\subfloat{\label{xfig:c}\includegraphics[width=.22\textwidth, height=1.5cm]{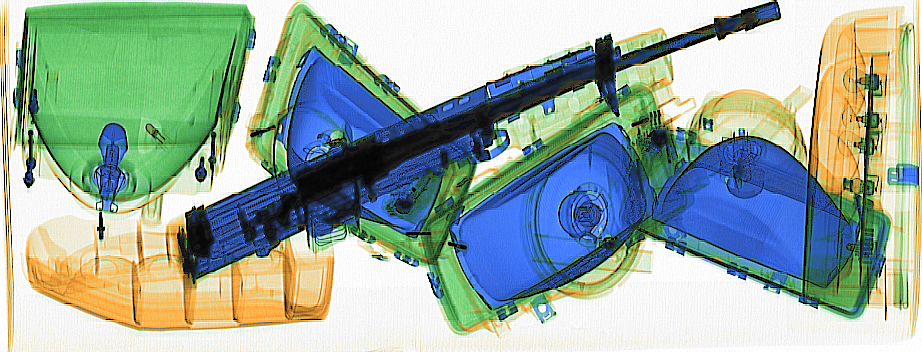}}\hfill
\subfloat{\label{xfig:d}\includegraphics[width=.22\textwidth, height=1.5cm]{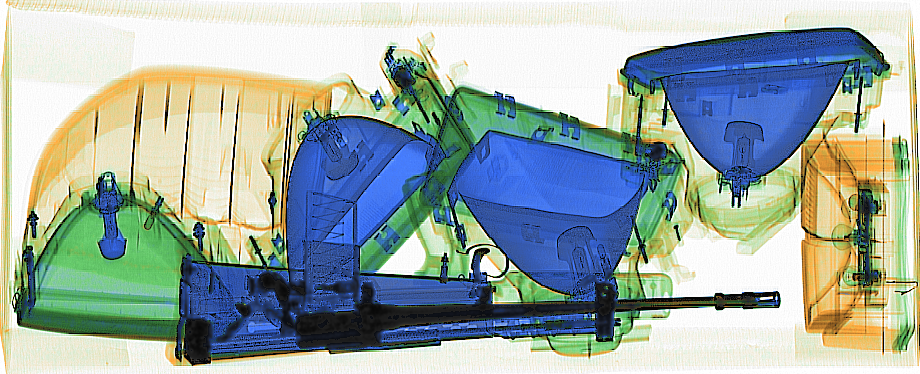}}\hfill    
\caption{Example dual-view images from the X-ray dataset. The top row is a SoC example and the bottom is a staged threat example.}
    \label{fig:xray_crops}
\end{figure}

\textbf{X-ray}: The X-ray dataset consists of 5,000 stream-of-commerce (SoC) parcels and 234 staged threat (threat) parcels collected from a UK parcel distribution centre. The SoC parcels consist of benign objects whilst the threat parcels also incorporate a firearm which is occasionally disassembled. All parcels are presented as a pair of dual-view images, which show the contents of the same parcel but are taken at perpendicular angles.  All images are false-coloured based on dual-energy imaging. Example images are shown in Figure \ref{fig:xray_crops}. 
The images were pre-processed in the same manner as described in \cite{griffin1}. Extra air was automatically removed and each image was split into $224 \times 224$ patches using a stride of 112 or less. The anomaly detectors used SoC patches for training and only encountered threat patches at the evaluation stage. 

\subsection{Knowledge Distillation}

\begin{figure}[h]
    \centering

\begin{tikzpicture}[x=0.75pt,y=0.75pt,yscale=-0.7,xscale=0.7]

\draw  [draw opacity=0][fill={rgb, 255:red, 155; green, 155; blue, 155 }  ,fill opacity=0.25 ] (130,29.48) .. controls (130,20.16) and (137.56,12.6) .. (146.88,12.6) -- (433.72,12.6) .. controls (443.04,12.6) and (450.6,20.16) .. (450.6,29.48) -- (450.6,80.12) .. controls (450.6,89.44) and (443.04,97) .. (433.72,97) -- (146.88,97) .. controls (137.56,97) and (130,89.44) .. (130,80.12) -- cycle ;
\draw (85,126) node  {\includegraphics[width=40pt,height=40pt]{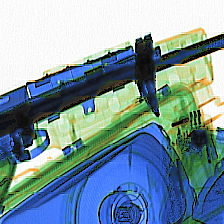}};
\draw   (141,34.68) .. controls (141,26.9) and (147.3,20.6) .. (155.08,20.6) -- (275.32,20.6) .. controls (283.1,20.6) and (289.4,26.9) .. (289.4,34.68) -- (289.4,76.92) .. controls (289.4,84.7) and (283.1,91) .. (275.32,91) -- (155.08,91) .. controls (147.3,91) and (141,84.7) .. (141,76.92) -- cycle ;
\draw   (141,173.96) .. controls (141,166.25) and (147.25,160) .. (154.96,160) -- (275.44,160) .. controls (283.15,160) and (289.4,166.25) .. (289.4,173.96) -- (289.4,215.84) .. controls (289.4,223.55) and (283.15,229.8) .. (275.44,229.8) -- (154.96,229.8) .. controls (147.25,229.8) and (141,223.55) .. (141,215.84) -- cycle ;
\draw    (85.4,55) -- (138,55) ;
\draw [shift={(141,55)}, rotate = 180] [fill={rgb, 255:red, 0; green, 0; blue, 0 }  ][line width=0.08]  [draw opacity=0] (10.72,-5.15) -- (0,0) -- (10.72,5.15) -- (7.12,0) -- cycle    ;
\draw    (85.4,55) -- (85.4,89.8) ;
\draw    (85.4,178.6) -- (85.4,195.6) ;
\draw    (85.4,195.6) -- (138.4,195.6) ;
\draw [shift={(141.4,195.6)}, rotate = 180] [fill={rgb, 255:red, 0; green, 0; blue, 0 }  ][line width=0.08]  [draw opacity=0] (10.72,-5.15) -- (0,0) -- (10.72,5.15) -- (7.12,0) -- cycle    ;
\draw   (311,25.28) .. controls (311,22.92) and (312.92,21) .. (315.28,21) -- (328.12,21) .. controls (330.48,21) and (332.4,22.92) .. (332.4,25.28) -- (332.4,85.72) .. controls (332.4,88.08) and (330.48,90) .. (328.12,90) -- (315.28,90) .. controls (312.92,90) and (311,88.08) .. (311,85.72) -- cycle ;
\draw   (311,165.28) .. controls (311,162.92) and (312.92,161) .. (315.28,161) -- (328.12,161) .. controls (330.48,161) and (332.4,162.92) .. (332.4,165.28) -- (332.4,225.72) .. controls (332.4,228.08) and (330.48,230) .. (328.12,230) -- (315.28,230) .. controls (312.92,230) and (311,228.08) .. (311,225.72) -- cycle ;
\draw    (288.67,55) -- (307.6,55) ;
\draw [shift={(310.6,55)}, rotate = 180] [fill={rgb, 255:red, 0; green, 0; blue, 0 }  ][line width=0.08]  [draw opacity=0] (10.72,-5.15) -- (0,0) -- (10.72,5.15) -- (7.12,0) -- cycle    ;
\draw    (289.4,194.8) -- (308.33,194.8) ;
\draw [shift={(311.33,194.8)}, rotate = 180] [fill={rgb, 255:red, 0; green, 0; blue, 0 }  ][line width=0.08]  [draw opacity=0] (10.72,-5.15) -- (0,0) -- (10.72,5.15) -- (7.12,0) -- cycle    ;
\draw [fill={rgb, 255:red, 142; green, 186; blue, 217 }  ,fill opacity=1 ]   (359.33,89.67) -- (441.33,89.67) ;
\draw  [color={rgb, 255:red, 0; green, 0; blue, 0 }  ,draw opacity=1 ][fill={rgb, 255:red, 142; green, 186; blue, 217 }  ,fill opacity=1 ] (359.33,49.67) -- (369.4,49.67) -- (369.4,89.67) -- (359.33,89.67) -- cycle ;
\draw  [fill={rgb, 255:red, 142; green, 186; blue, 217 }  ,fill opacity=1 ] (369.4,39.8) -- (379.47,39.8) -- (379.47,89.67) -- (369.4,89.67) -- cycle ;
\draw  [fill={rgb, 255:red, 142; green, 186; blue, 217 }  ,fill opacity=1 ] (379.33,59.8) -- (389.4,59.8) -- (389.4,89.67) -- (379.33,89.67) -- cycle ;
\draw  [fill={rgb, 255:red, 142; green, 186; blue, 217 }  ,fill opacity=1 ] (389.27,49.67) -- (399.33,49.67) -- (399.33,89.67) -- (389.27,89.67) -- cycle ;
\draw  [fill={rgb, 255:red, 142; green, 186; blue, 217 }  ,fill opacity=1 ] (399.53,29.8) -- (409.6,29.8) -- (409.6,89.67) -- (399.53,89.67) -- cycle ;
\draw  [fill={rgb, 255:red, 142; green, 186; blue, 217 }  ,fill opacity=1 ] (409.53,41) -- (419.6,41) -- (419.6,89.27) -- (409.53,89.27) -- cycle ;
\draw  [fill={rgb, 255:red, 142; green, 186; blue, 217 }  ,fill opacity=1 ] (419.33,65) -- (429.4,65) -- (429.4,89.67) -- (419.33,89.67) -- cycle ;
\draw  [fill={rgb, 255:red, 142; green, 186; blue, 217 }  ,fill opacity=1 ] (429.4,49.67) -- (439.47,49.67) -- (439.47,89.67) -- (429.4,89.67) -- cycle ;
\draw [fill={rgb, 255:red, 255; green, 190; blue, 134 }  ,fill opacity=1 ]   (360.33,229.67) -- (442.33,229.67) ;
\draw  [fill={rgb, 255:red, 255; green, 190; blue, 134 }  ,fill opacity=1 ] (360.33,194.6) -- (370.4,194.6) -- (370.4,229.67) -- (360.33,229.67) -- cycle ;
\draw  [fill={rgb, 255:red, 255; green, 190; blue, 134 }  ,fill opacity=1 ] (370.4,179.8) -- (380.47,179.8) -- (380.47,229.67) -- (370.4,229.67) -- cycle ;
\draw  [fill={rgb, 255:red, 255; green, 190; blue, 134 }  ,fill opacity=1 ] (380.33,206.6) -- (390.4,206.6) -- (390.4,229.67) -- (380.33,229.67) -- cycle ;
\draw  [fill={rgb, 255:red, 255; green, 190; blue, 134 }  ,fill opacity=1 ] (390.27,159.8) -- (400.33,159.8) -- (400.33,229.67) -- (390.27,229.67) -- cycle ;
\draw  [fill={rgb, 255:red, 255; green, 190; blue, 134 }  ,fill opacity=1 ] (400.53,225.8) -- (410.6,225.8) -- (410.6,229.67) -- (400.53,229.67) -- cycle ;
\draw  [fill={rgb, 255:red, 255; green, 190; blue, 134 }  ,fill opacity=1 ] (410.53,210.8) -- (420.6,210.8) -- (420.6,229.27) -- (410.53,229.27) -- cycle ;
\draw  [fill={rgb, 255:red, 255; green, 190; blue, 134 }  ,fill opacity=1 ] (420.33,181.8) -- (430.4,181.8) -- (430.4,229.67) -- (420.33,229.67) -- cycle ;
\draw  [fill={rgb, 255:red, 255; green, 190; blue, 134 }  ,fill opacity=1 ] (430.4,189.67) -- (440.47,189.67) -- (440.47,229.67) -- (430.4,229.67) -- cycle ;
\draw    (400.55,102.8) -- (400.2,122.6) ;
\draw [shift={(400.6,99.8)}, rotate = 91.01] [fill={rgb, 255:red, 0; green, 0; blue, 0 }  ][line width=0.08]  [draw opacity=0] (10.72,-5.15) -- (0,0) -- (10.72,5.15) -- (7.12,0) -- cycle    ;
\draw [fill={rgb, 255:red, 142; green, 186; blue, 217 }  ,fill opacity=1 ]   (335.47,55) -- (354.4,55) ;
\draw [shift={(357.4,55)}, rotate = 180] [fill={rgb, 255:red, 0; green, 0; blue, 0 }  ][line width=0.08]  [draw opacity=0] (10.72,-5.15) -- (0,0) -- (10.72,5.15) -- (7.12,0) -- cycle    ;
\draw [fill={rgb, 255:red, 255; green, 190; blue, 134 }  ,fill opacity=1 ]   (334.87,195) -- (353.8,195) ;
\draw [shift={(356.8,195)}, rotate = 180] [fill={rgb, 255:red, 0; green, 0; blue, 0 }  ][line width=0.08]  [draw opacity=0] (10.72,-5.15) -- (0,0) -- (10.72,5.15) -- (7.12,0) -- cycle    ;
\draw    (400.2,134.6) -- (400.2,145.4) ;
\draw  [draw opacity=0][fill={rgb, 255:red, 155; green, 155; blue, 155 }  ,fill opacity=0.1 ] (130,170.08) .. controls (130,160.76) and (137.56,153.2) .. (146.88,153.2) -- (433.72,153.2) .. controls (443.04,153.2) and (450.6,160.76) .. (450.6,170.08) -- (450.6,220.72) .. controls (450.6,230.04) and (443.04,237.6) .. (433.72,237.6) -- (146.88,237.6) .. controls (137.56,237.6) and (130,230.04) .. (130,220.72) -- cycle ;

\draw (52,165) node [anchor=north west][inner sep=0.75pt]  [font=\scriptsize] [align=center] {Input image};
\draw (400.2,128.5) [font=\scriptsize] [align=center ]node {{Compare difference}};
\draw (170,50) node [anchor=north west][inner sep=0.75pt]  [font=\scriptsize] [align=center] {Teacher network};
\draw (170,190) node [anchor=north west][inner sep=0.75pt]  [font=\scriptsize] [align=center] {Student network};
\draw (316.33,84.78) node [anchor=north west][inner sep=0.75pt]  [font=\scriptsize,rotate=-270] [align=center] {Projection};
\draw (316.33,224.78) node [anchor=north west][inner sep=0.75pt]  [font=\scriptsize,rotate=-270] [align=center] {Projection};

\end{tikzpicture}

    \caption{Schematic of the knowledge distillation architecture.}
    \label{fig:kd}
\end{figure}
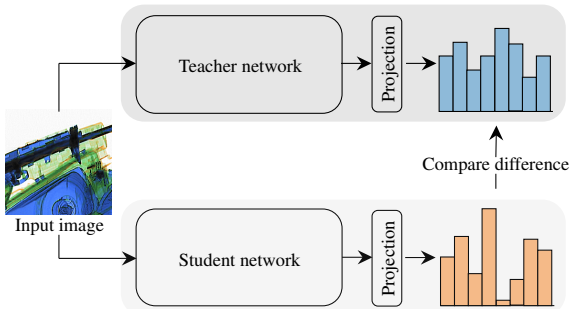

We used knowledge distillation (KD) as the primary anomaly detector. Figure \ref{fig:kd} depicts a schematic of the architecture. At training time, an image from the normal class is fed into both the student and teacher. The teacher is pre-trained on an auxiliary task (except in the case of randomly initialised weights). The teacher's weights are frozen in the distillation stage and the student is trained to match the representations output by the teacher. We append a linear projection head to the final pooling layer and treat this as the output. This was done to ensure fairer comparisons between tasks. The student is optimised using MSE. 

MSE between the student and teacher, evaluated on a test datum, is used as the anomaly score. We found that anomaly detection performance was not improved by ensembling or by using more sophisticated scoring methods. For an input datum $x$, let the output of the student be $f(x)$ and the output of the teacher be $h(x)$. The anomaly score $A(x)$ is then defined as:

\begin{equation}
    A(x) = ||h(x) - f(x)||^2
\end{equation}

The student and teacher have identical architectures for all experiments. This was done to minimise architecture search but this is not a requirement for the method. As it was found that image augmentations could erode features that distinguished normal data from anomalies, no augmentations were applied to the input data. All students were trained with an Adam optimiser with a learning rate of 1e-5 for 20 epochs, as it was found that larger learning rates lead to more training instability and overshooting. 

\textbf{Why does knowledge distillation work as an anomaly detector?} The student should output similar representations as the teacher for normal data because it has encountered similar examples during training. This is reliant on the assumption that the normal data's distribution distilled from the teacher is powerful enough that it generalises well to unseen normal instances. In contrast, the student's representations for anomalies should differ compared to the teacher, as the student cannot extrapolate and it is expected the teacher learns more diverse features due to the wider scope of the auxiliary task. 

\textbf{How were the teacher representations selected?} We used random weights, transferred representations from supervised classification tasks, and trained representations from scratch using self-supervised methods \cite{simclr, rotnet, Vincent_Larochelle_Bengio_Manzagol_2008}. For auxiliary tasks that used the anomaly detection dataset, we used the normal data which was also used to train the student in the knowledge distillation stage, and the anomalous class was never included in pre-training.

The specific auxiliary tasks for each anomaly detection dataset are described in Table \ref{fig:tab1} and further detail about the datasets and tasks are described in the appendix. The network architectures for each anomaly detection dataset are as follows:

\begin{table*}
\centering
\caption{Auxiliary tasks and datasets used to pre-train the teacher and generate fixed representations for each anomaly detection dataset, excluding randomly initialised weights which are used for all datasets. Baseline representations are italicised.}
\label{fig:tab1}
\footnotesize
\begin{tabular}{llllll} 
\toprule
                                                         & \multicolumn{5}{c}{\textbf{Auxiliary Task} }                                                                                                                                                                                                                                                                                                                                                                                                                                                                                                                                                                                                                                                                                                                \\ 
\cmidrule{2-6}
\textbf{Dataset}                                         & Classification                                                                                                                                                                 & RotNet                                                                                                                                                 & Autoencoder                                                                                                                                            & \begin{tabular}[c]{@{}l@{}}Denoising \\Autoencoder \end{tabular}                                                                                       & SimCLR                                                                                          \\ 
\midrule
Cats vs. Dogs                                               & \begin{tabular}[c]{@{}l@{}}\begin{tabular}{@{\labelitemi\hspace{\dimexpr\labelsep+0.5\tabcolsep}}l}STL\\CIFAR-10\\\textit{Cats vs. Dogs}\end{tabular}\end{tabular}                & \begin{tabular}[c]{@{}l@{}}\begin{tabular}{@{\labelitemi\hspace{\dimexpr\labelsep+0.5\tabcolsep}}l}STL\\CIFAR-10\\Cats vs. Dogs\end{tabular}\end{tabular} & \begin{tabular}[c]{@{}l@{}}\begin{tabular}{@{\labelitemi\hspace{\dimexpr\labelsep+0.5\tabcolsep}}l}STL\\CIFAR-10\\Cats vs. Dogs\end{tabular}\end{tabular} & \begin{tabular}[c]{@{}l@{}}\begin{tabular}{@{\labelitemi\hspace{\dimexpr\labelsep+0.5\tabcolsep}}l}STL\\CIFAR-10\\Cats vs. Dogs\end{tabular}\end{tabular} & \begin{tabular}{@{\labelitemi\hspace{\dimexpr\labelsep+0.5\tabcolsep}}l}Not used \end{tabular}  \\ 
\midrule
CIFAR-10                                                 & \begin{tabular}[c]{@{}l@{}}\begin{tabular}{@{\labelitemi\hspace{\dimexpr\labelsep+0.5\tabcolsep}}l}STL\\FMNIST \\\textit{CIFAR-10 }\end{tabular}\end{tabular}                  & \begin{tabular}[c]{@{}l@{}}\begin{tabular}{@{\labelitemi\hspace{\dimexpr\labelsep+0.5\tabcolsep}}l}STL\\FMNIST\\CIFAR-10 \end{tabular}\end{tabular}    & \begin{tabular}[c]{@{}l@{}}\begin{tabular}{@{\labelitemi\hspace{\dimexpr\labelsep+0.5\tabcolsep}}l}STL\\FMNIST\\CIFAR-10 \end{tabular}\end{tabular}    & \begin{tabular}[c]{@{}l@{}}\begin{tabular}{@{\labelitemi\hspace{\dimexpr\labelsep+0.5\tabcolsep}}l}STL\\FMNIST\\CIFAR-10 \end{tabular}\end{tabular}    & \begin{tabular}{@{\labelitemi\hspace{\dimexpr\labelsep+0.5\tabcolsep}}l}CIFAR-10 \end{tabular}  \\ 
\midrule
\begin{tabular}[c]{@{}l@{}}Plant\\Pathology\end{tabular} & \begin{tabular}[c]{@{}l@{}}\begin{tabular}{@{\labelitemi\hspace{\dimexpr\labelsep+0.5\tabcolsep}}l}Plant Village \cite{hughes2016open}\\ImageNet\\\textit{Plant Pathology}\end{tabular}\end{tabular} & \begin{tabular}{@{\labelitemi\hspace{\dimexpr\labelsep+0.5\tabcolsep}}l}Plant Pathology\end{tabular}                                                   & \begin{tabular}{@{\labelitemi\hspace{\dimexpr\labelsep+0.5\tabcolsep}}l}Not used\end{tabular}                                                          & \begin{tabular}{@{\labelitemi\hspace{\dimexpr\labelsep+0.5\tabcolsep}}l}Not used\end{tabular}                                                          & \begin{tabular}{@{\labelitemi\hspace{\dimexpr\labelsep+0.5\tabcolsep}}l}Not used \end{tabular}  \\ 
\midrule
X-ray                                                    & \begin{tabular}[c]{@{}l@{}}\begin{tabular}{@{\labelitemi\hspace{\dimexpr\labelsep+0.5\tabcolsep}}l}ImageNet \\\textit{X-ray}\end{tabular}\end{tabular}                         & \begin{tabular}{@{\labelitemi\hspace{\dimexpr\labelsep+0.5\tabcolsep}}l}X-ray \end{tabular}                                                            & \begin{tabular}{@{\labelitemi\hspace{\dimexpr\labelsep+0.5\tabcolsep}}l}Not used \end{tabular}                                                         & \begin{tabular}{@{\labelitemi\hspace{\dimexpr\labelsep+0.5\tabcolsep}}l}Not used \end{tabular}                                                         & \begin{tabular}{@{\labelitemi\hspace{\dimexpr\labelsep+0.5\tabcolsep}}l}Not used \end{tabular}  \\
\bottomrule
\end{tabular}
\end{table*}

\textbf{Cats vs. Dogs and CIFAR-10}: A small ResNet \cite{cifar10fast} was used for all auxiliary tasks. The classification head was removed and a 128-dimensional randomly initialised head was appended to the final pooling layer. Correspondingly, this architecture was mirrored for the autoencoders. The autoencoders were trained with a 128-dimensional bottleneck and the decoder was discarded during knowledge distillation. Note for SimCLR we use the same transformations to pre-train the network as the original paper \cite{simclr}.

\textbf{Plant Pathology and X-ray}: We extracted the representations after the final pooling layer of a DenseNet-161 network \cite{huang2017densely}. For the X-ray task, as patches were fed into the network instead of full images, the maximum MSE across an image's patches was used as the anomaly score. 

\subsection{Baseline Comparisons}

For all four datasets, we defined a ceiling on anomaly detection performance by conducting knowledge distillation using a teacher representation which was pre-trained to discriminate between normal data and anomalies in a supervised manner. These are referred to as the baseline representations. 

We also recorded the supervised classification performance baseline by freezing the weights of the networks trained on the auxiliary tasks, appending a linear head, and fine-tuning the networks. The results are reported as classification accuracies. For CIFAR-10, the networks were fine-tuned to classify between all ten classes whilst for the other three datasets the networks were fine-tuned to distinguish between the two classes.

Finally, we compared the performance of knowledge distillation as an anomaly detector by using MSE and Mahalanobis distance \cite{lee_mahalanobis} on the teacher representations directly as alternative anomaly detector methods. More concretely, we computed the mean and covariance of normal training data using the teacher representations. We then used the MSE and Mahalanobis distance between a test datum in the teacher representation space and these values to produce anomaly scores. The MSE results are reported in the appendix.

\subsection{Evaluating Anomaly Detection Performance}
We evaluate the performance of the anomaly detectors at different thresholds using the Area Under the Receiving Operator Curve (AUROC). AUROC can be viewed as the probability that an out-of-distribution sample is given a higher anomaly score than an in-distribution sample. Hence, a score of 50\% indicates the detector performs at random whilst a score of 100\% indicates a perfect anomaly detector. AUROC is preferred over other measurements such as accuracy due to the class imbalance between normal and anomalous instances. 

\subsection{Measuring Brittleness}
After training a student through knowledge distillation, we modify the procedure of Simon-Gabriel \etal. \cite{pmlr-v97-simon-gabriel19a} and record the average L2 gradient norms of the normal training data with respect to the student network. We amend the computation by dividing the average L2 gradient norms by the trace of the covariance data to account for the spread of different representations and to allow for meaningful comparisons of this score across representations:

\begin{equation}
\label{eqn:l2norm}
\frac{\mathbb{E}||{{\partial}_x}_{\mathrm{train}} L||_2}{\mathrm{tr}(\Sigma_{\mathrm{train}})}
\end{equation}

Note that $x_\mathrm{train}$ denotes a training sample in the normal dataset, $L$ is the MSE loss function (between the student and teacher outputs) and $\mathrm{tr}(\Sigma_{\mathrm{train}})$ denotes the trace of the training data's covariance matrix, which is computed using the difference between the student and teacher outputs for each training datum.

More adversarially vulnerable instances require smaller shifts in the input domain to evoke a change in the model's output prediction. Hence, a larger gradient norm is indicative of increased susceptibility to adversarial perturbations.
\section{Results}

Subsets of the anomaly detection results are reported in Tables \ref{fig:cvdres}, \ref{fig:cifarres}, \ref{fig:plantres} and \ref{fig:xrayres}. The complete set of results can be found in the appendix. For each dataset, the better anomaly detection performance between the two methods is indicated in bold.

\begin{table}
\centering
\caption{Cats vs. Dogs results averaged over the two classes.}
   \label{fig:cvdres}
   \footnotesize
\begin{tabular}{llll} 
\toprule
                                                                                  &                                                                                & \multicolumn{2}{c}{\begin{tabular}[c]{@{}c@{}}\textbf{Anomaly Detection}\\\textbf{Method (AUROC) }\end{tabular}}  \\ 
\cmidrule{3-4}
\begin{tabular}[c]{@{}l@{}}\textbf{Teacher}\\\textbf{Representation}\end{tabular} & \begin{tabular}[c]{@{}l@{}}\textbf{Classification}\\\textbf{Acc.}\end{tabular} & \begin{tabular}[c]{@{}l@{}}Knowledge\\Distillation\end{tabular} & Mahalanobis                                     \\ 
\midrule
\textit{Baseline}                                                                 & \textit{83.40}                                                                 & \textit{\textbf{89.37}}                                         & \textit{85.93}                                  \\ 
\midrule
Random                                                                            & 64.11                                                                          & 50.87                                                           & \textbf{51.17}                                  \\ 
\midrule
STL Class.                                                                        & 71.28                                                                          & \textbf{61.82}                                                  & 56.34                                           \\
CIFAR Class.                                                                      & 87.57                                                                          & \textbf{81.98}                                                  & 74.91                                           \\ 
\midrule
STL RotNet                                                                        & 75.87                                                                          & \textbf{54.91}                                                  & 53.96                                           \\
CIFAR RotNet                                                                      & 76.65                                                                          & \textbf{56.69}                                                  & 55.85                                           \\
CvD RotNet                                                                        & 70.12                                                                          & \textbf{50.18}                                                  & 49.64                                           \\ 
\midrule
STL AE                                                                            & 64.66                                                                          & \textbf{52.03}                                                  & 51.77                                           \\
CIFAR AE                                                                          & 64.11                                                                          & \textbf{51.63}                                                  & 51.10                                           \\
CvD AE                                                                            & 64.47                                                                          & \textbf{51.29}                                                  & 50.63                                           \\ 
\midrule
STL DAE                                                                           & 64.69                                                                          & \textbf{53.65}                                                  & 51.52                                           \\
CIFAR DAE                                                                         & 66.13                                                                          & \textbf{53.42}                                                  & 51.50                                           \\
CvD DAE                                                                           & 57.23                                                                          & \textbf{50.87}                                                  & 50.66                                           \\
\bottomrule
\end{tabular}
\end{table}

\begin{table}
\centering
\caption{CIFAR-10 results averaged over unimodal and multimodal configurations. Separate unimodal and multimodal results are located in the appendix.}
   \label{fig:cifarres}
   \footnotesize
\begin{tabular}{llll} 
\toprule
                                                                                  &                                                                                & \multicolumn{2}{c}{\begin{tabular}[c]{@{}c@{}}\textbf{Anomaly Detection}\\\textbf{Method (AUROC)}\end{tabular}}  \\
\cmidrule{3-4}
\begin{tabular}[c]{@{}l@{}}\textbf{Teacher}\\\textbf{Representation}\end{tabular} & \begin{tabular}[c]{@{}l@{}}\textbf{Classification}\\\textbf{Acc.}\end{tabular} & \begin{tabular}[c]{@{}l@{}}Knowledge\\Distillation\end{tabular} & Mahalanobis                                    \\ 
\midrule
\textit{Baseline}                                                                 & \textit{94.08}                                                                 & \textit{\textbf{91.52}}                                         & \textit{79.77}                                 \\ 
\midrule
Random                                                                            & 40.31                                                                          & \textbf{55.53}                                                  & 54.19                                          \\ 
\midrule
STL Class.                                                                        & 56.70                                                                          & \textbf{80.78}                                                  & 73.00                                          \\
FMNIST Class.                                                                     & 26.16                                                                          & \textbf{54.34}                                                  & 53.70                                          \\ 
\midrule
STL RotNet                                                                        & 59.94                                                                          & \textbf{63.01}                                                  & 59.61                                          \\
FMNIST RotNet                                                                     & 21.59                                                                          & \textbf{54.68}                                                  & 51.67                                          \\
CIFAR RotNet                                                                      & 37.27                                                                          & \textbf{73.35}                                                  & 69.23                                          \\ 
\midrule
CIFAR SimCLR                                                                      & 65.04                                                                          & \textbf{51.78}                                                  & 47.15                                          \\ 
\midrule
STL AE                                                                            & 47.13                                                                          & \textbf{57.16}                                                  & 56.29                                          \\
FMNIST AE                                                                         & 43.60                                                                          & \textbf{57.69}                                                  & 55.26                                          \\
CIFAR AE                                                                          & 44.50                                                                          & \textbf{56.33}                                                  & 55.28                                          \\ 
\midrule
STL DAE                                                                           & 49.82                                                                          & 55.73                                                           & \textbf{57.68}                                 \\
FMNIST DAE                                                                        & 40.57                                                                          & \textbf{56.25}                                                  & 54.26                                          \\
CIFAR DAE                                                                         & 14.67                                                                          & \textbf{55.45}                                                  & 54.75                                          \\
\bottomrule
\end{tabular}
\end{table}

\begin{table}
\centering
\caption{Plant Pathology results.}
\label{fig:plantres}
\footnotesize
\begin{tabular}{llll} 
\toprule
\multicolumn{1}{c}{}                                                              &                                                                                & \multicolumn{2}{c}{\textbf{Anomaly Detection Method }}                            \\ 
\cmidrule{3-4}
\begin{tabular}[c]{@{}l@{}}\textbf{Teacher}\\\textbf{Representation}\end{tabular} & \begin{tabular}[c]{@{}l@{}}\textbf{Classification}\\\textbf{Acc.}\end{tabular} & \begin{tabular}[c]{@{}l@{}}Knowledge\\Distillation\end{tabular} & Mahalanobis     \\ 
\midrule
\textit{Baseline }                                                                & \textit{100}                                                                   & \textit{\textbf{100}}                                           & \textit{99.89}  \\ 
\midrule
Random                                                                            & 63.67                                                                          & \textbf{42.50}                                                  & 41.35           \\ 
\midrule
Plant Village Class.                                                              & 90.92                                                                          & 89.82                                                           & \textbf{90.78}  \\
ImageNet Class.                                                                   & 88.24                                                                          & \textbf{70.66}                                                  & 49.45           \\ 
\midrule
Plant Path. RotNet                                                                & 57.97                                                                          & \textbf{48.09}                                                  & 47.51           \\
\bottomrule
\end{tabular}
\end{table}

\begin{table}
\centering
\caption{X-ray results.}
\label{fig:xrayres}
\footnotesize
\begin{tabular}{llll} 
\toprule
\multicolumn{1}{c}{}                                                              &                                                                                & \multicolumn{2}{c}{\textbf{Anomaly Detection Method }}                             \\ 
\cmidrule{3-4}
\begin{tabular}[c]{@{}l@{}}\textbf{Teacher}\\\textbf{Representation}\end{tabular} & \begin{tabular}[c]{@{}l@{}}\textbf{Classification}\\\textbf{Acc.}\end{tabular} & \begin{tabular}[c]{@{}l@{}}Knowledge\\Distillation\end{tabular} & Mahalanobis      \\ 
\midrule
\textit{Baseline }                                                                & \textit{98.98 }                                                                & \textbf{\textit{99.74}}                                         & \textit{99.63 }  \\ 
\midrule
Random                                                                            & 88.56                                                                          & \textbf{73.61}                                                  & 71.87            \\ 
\midrule
ImageNet Class.                                                                   & 96.79                                                                          & \textbf{76.36}                                                  & 73.46            \\ 
\midrule
X-ray RotNet                                                                      & 93.62                                                                          & \textbf{96.41}                                                  & 60.06            \\
\bottomrule
\end{tabular}
\end{table}


\textbf{Anomaly detection performance is more dependent on the representation than the anomaly detector.} Although knowledge distillation generally outperforms alternate anomaly detection methods, its rate of success depends on the underlying teacher representation. 

\begin{figure}[h]
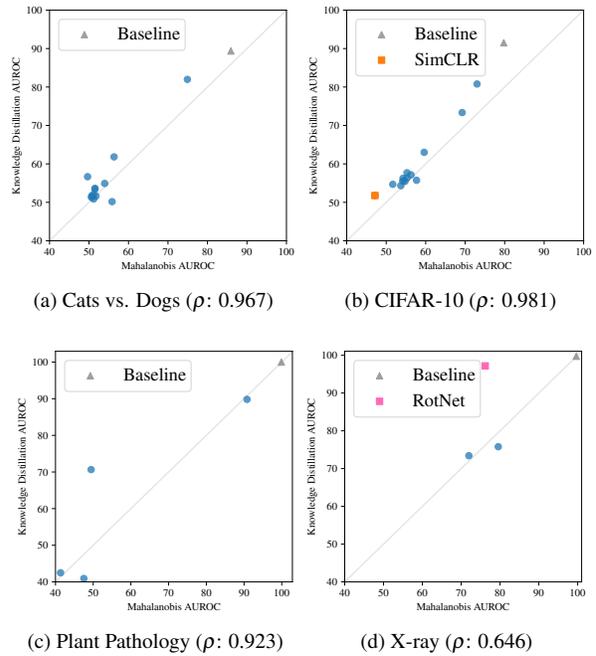

    \centering
\subfloat[Cats vs. Dogs ($\rho$: 0.967)]{\label{sfig:cvdmahkd}\scalebox{0.4}{\input{graphs/cvd_mahal_kd2.pgf}}}
\subfloat[CIFAR-10 ($\rho$: 0.981)]{\label{sfig:cifarmahkd}\scalebox{0.4}{\input{graphs/cifar_mahal_kd2.pgf}}}\hfill
\subfloat[Plant Pathology ($\rho$: 0.923)]{\label{sfig:plantmahkd}\scalebox{0.4}{\input{graphs/plant_mahal_kd2.pgf}}}
\subfloat[X-ray ($\rho$: 0.646)]{\label{sfig:xmahkd}\scalebox{0.4}{\input{graphs/xray_mahal_kd2.pgf}}}\hfill
\caption{{Scatter plots of Mahalanobis AUROC scores against knowledge distillation AUROC scores for each representation.}}
    \label{fig:mahkd}
\end{figure}

The scatter plots in Figure \ref{fig:mahkd} illustrate the relationship between knowledge distillation and Mahalanobis performance across all representations. There is a clear correlation between these scores for all datasets (as indicated by the correlation coefficient $\rho$ which is reported for each dataset), reinforcing that the choice of representation is more important than the anomaly detector. 

The more successful teacher representations are a result of the teacher learning from classes that resemble both anomalous and normal classes at pre-training. For example, CIFAR-10 classification works well for Cats vs. Dogs as it includes cat and dog classes and Plant Village includes examples of both healthy and diseased leaves. This represents a clear imbalance between the knowledge of the teacher and the student, as the student does not encounter anomalies during training. In addition, the observation that classification on STL is the best teacher representation for CIFAR-10 suggests that the size of the pre-training set is not as important as the similarity of classes encountered during training, as STL is a much smaller size.

\textbf{Knowledge distillation outperforms other methods in multimodal normal settings.} Observing the relationship between knowledge distillation and Mahalanobis performance on CIFAR-10 in Figure \ref{fig:cifarunimulti}, there is more benefit to using knowledge distillation in the multimodal setting compared to the unimodal setting.

\begin{figure}[h]
    \centering
    \scalebox{0.6}{\input{graphs/cifarunimulti.pgf}}
    \caption*{Unimodal $\rho$: 0.985; Multimodal $\rho$: 0.953}
    \caption{Scatter plot of CIFAR-10 Mahalanobis AUROC scores against knowledge distillation AUROC scores, separated by unimodal and multimodal configurations.}
    \label{fig:cifarunimulti}
\end{figure}

Use of the Mahalanobis score and MSE directly on the representations relies on the assumption that the normal data is modelled by a multivariate Gaussian distribution. This may be reasonable for a unimodal setting where the normal class is tighter, but not for the multimodal setting, where there is more intra-class variation. 

In contrast, knowledge distillation utilises the errors between student and teacher representations as opposed to using the teacher representations directly. Hence, there are no such constraints on the distribution of the normal class. As the detector is trained through a regression task, at most, one can assume that the errors are iid and Gaussian distributed. In turn, this motivates the use of MSE as an anomaly score in the knowledge distillation configuration. 

\subsection{Connections with Separability}
\begin{figure}[h]
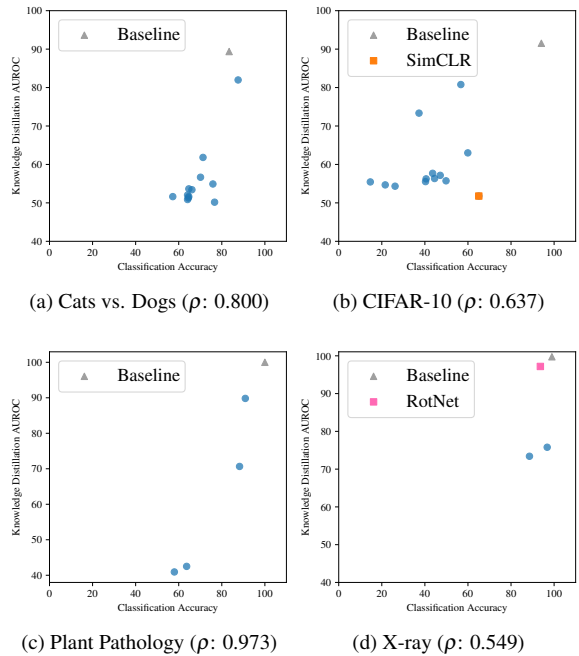

    \centering
\subfloat[Cats vs. Dogs ($\rho$: 0.800)]{\label{sfig:cvdsupkd}\scalebox{0.4}{\input{graphs/cvd_kdacc.pgf}}}
\subfloat[CIFAR-10 ($\rho$: 0.637)]{\label{sfig:cifarsupkd}\scalebox{0.4}{\input{graphs/cifar_kdacc.pgf}}}\hfill
\subfloat[Plant Pathology ($\rho$: 0.973)]{\label{sfig:plantsupkd}\scalebox{0.4}{\input{graphs/plant_kdacc.pgf}}}
\subfloat[X-ray ($\rho$: 0.549)]{\label{sfig:xsupkd}\scalebox{0.4}{\input{graphs/xray_kdacc.pgf}}}\hfill
\caption{Scatter plots of supervised classification accuracies against knowledge distillation AUROC scores for each representation.}
    \label{fig:supkd}
\end{figure}

\textbf{Separability between anomalies and normal data is not the sole factor for good anomaly detection.} We investigate the role of representations further by plotting the relationship between knowledge distillation and supervised classification accuracy as shown in Figure \ref{fig:supkd}. There is a correlation between these values but not an exact one-to-one correspondence. 

On CIFAR-10, although SimCLR achieves the highest supervised classification performance other than the baseline, its anomaly detection performance is underwhelming. Additionally, although RotNet does not have the highest classification performance amongst the X-ray representations, its knowledge distillation performance vastly outperforms the others, as well as the previous anomaly detection performance of 92.5\% recorded in \cite{griffin1}. Both representations have been highlighted in their respective subplots.  We discuss a possible reason for this in subsection \ref{sec:non-robust}.

\subsection{Connections with Brittle Representations}
\label{sec:non-robust}

Although the anomaly detection results highlight a link with supervised classification performance, supervised performance does not appear to be the sole factor for good anomaly detection. We hypothesise that anomaly detectors also rely on non-robust features. Namely, the distribution of normal data in a good representation space has directions in which the distribution is tight, hence the features are `brittle'. Consequently, these brittle features provide an opportunity for anomalous data to manifest its unusualness. 

\begin{figure}[h!]
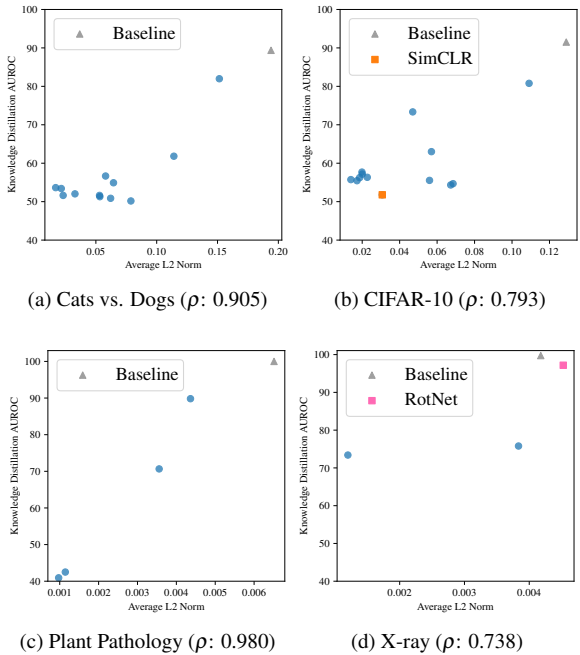

    \centering
\subfloat[Cats vs. Dogs ($\rho$: 0.905)]{\label{sfig:cvdl2kd}\scalebox{0.4}{\input{graphs/cvd_kdl2.pgf}}}
\subfloat[CIFAR-10 ($\rho$: 0.793)]{\label{sfig:cifarl2kd}\scalebox{0.4}{\input{graphs/cifar_kdl2.pgf}}}\hfill
\subfloat[Plant Pathology ($\rho$: 0.980)]{\label{sfig:plantl2kd}\scalebox{0.4}{\input{graphs/plant_kdl2.pgf}}}
\subfloat[X-ray ($\rho$: 0.738)]{\label{sfig:xl2kd}\scalebox{0.4}{\input{graphs/xray_kdl2.pgf}}}\hfill
\caption{Scatter plots of average L2 gradient norms against knowledge distillation AUROC scores for each representation.}
    \label{fig:l2kd}
\end{figure}

\textbf{Better representations for anomaly detection are correlated with larger average gradient norms.} We illustrate the relationship between anomaly detection performance and brittle features by plotting the relationship between the average L2 gradient norms (see Equation \ref{eqn:l2norm}), and knowledge distillation performance in Figure \ref{fig:l2kd}. Generally, there is a positive correlation between better performance and larger gradient norms. Note that this could provide an explanation for SimCLR's performance on CIFAR-10 (as its average L2 norm is lower compared to other representations) and RotNet's performance on X-ray (as its average L2 norm is large, and is actually close to the baseline's value). This suggests larger norms could be used to indicate which representations are better candidates for the anomaly detection task. 

\section{Discussion}
We showed that the success of an anomaly detector is highly dependent on the underlying representation. If one cannot test supervised classification performance through the use of proxy anomalies to get a diagnostic for how well normal data and anomalies are separated, our results suggest a good representation could be obtained by pre-training a neural network to classify on a similar dataset, or by pre-training RotNet on the normal data (if orientation is a salient feature for the normal data). Larger L2 gradient norms could also be used as another indicator for better representations.
\section{Conclusion}

We evaluated the performance of knowledge distillation as an anomaly detector and also the effects of different representations on anomaly detection performance. We showed that the choice of representation is more important than the anomaly detector, although evaluating errors is better than using the representation directly in the case of more complex normal data. Representations that are better suited for anomaly detection provide strong hints about how normal data and anomalies are separated, and there is a connection between improved anomaly detection performance and the use of brittle features to characterise normal data. Finally, we outperformed the previous anomaly detection performance results on an X-ray anomaly detection dataset, increasing performance from 92.8\% to 96.4\%. 
\section*{Acknowledgements}

This work was supported by the Engineering and Physical Sciences Research Council (EPSRC) under grant EP/R513143/1. Kimberly T. Mai also thanks Kelvin Ma and the anonymous reviewers for their feedback on this paper.

{\small
\bibliographystyle{ieee_fullname}
\bibliography{ktbib}
}

\begin{appendices}
\section{Appendix}

\subsection{Auxiliary Task Details}
Classification networks were pre-trained to distinguish between all of the provided annotated labels using a cross-entropy loss.

In the RotNet task \cite{rotnet}, each input image is rotated a multiple of $90$ degrees ($[0^{\circ}, 90^{\circ}, 180^{\circ}, 270^{\circ}]$) and the model is tasked with predicting the extent of rotation. The networks were trained using a cross-entropy loss. 

Autoencoders were tasked with reconstructing the input image and were optimised using a mean squared error loss.

For denoising autoencoders, Gaussian noise was added to each input image and the denoising autoencoder was trained to remove this corruption. Denoising autoencoders were also optimised using a mean squared error loss.

\subsection{Additional Datasets}
STL-10 (STL) \cite{pmlr-v15-coates11a} is an image dataset inspired by CIFAR-10. There are ten classes in the dataset (500 training images and 800 testing images per class) and each image is $96 \times 96$. The images are also supplemented by 100,000 for unlabelled images for unsupervised training. For pre-training, each image was resized to $32 \times 32$. For the classification pre-training task, only labelled data was used. Unlabelled data was incorporated in the training of the RotNets and autoencoders.

Fashion MNIST (FMNIST) \cite{xiao2017/online} is a greyscale dataset containing images from ten different clothing classes. There are 60,000 training images and 10,000 testing images in total, each of size $28 \times 28$. For pre-training, each image was resized to $32 \times 32$.

Plant Village \cite{hughes2016open} consists of 54,309 healthy and unhealthy leaf images spanning 14 crop species. The images are further categorised into healthy or containing a disease, leading to 39 disjoint classes in total.

ImageNet (ILSVRC2012) \cite{ILSVRC15} consists of 1.2 million training photographs segmented into 1,000 classes.

\subsection{Additional Anomaly Detection Methods}
In addition to knowledge distillation and the complete Mahalanobis distance, we also used the mean squared error (MSE) distance and a diagonalised Mahalanobis to compute anomaly detection scores. MSE denotes the mean squared error from the mean training representation whilst diagonalised Mahalanobis uses a diagonalised covariance of the training representation instead of the full covariance.

\begin{table*}[h]
\centering
\caption{Anomaly detection results for the cat class in the Cats vs. Dogs dataset.}
\label{fig:cvdrescat}
\footnotesize
\begin{tabular}{lllllll} 
\toprule
                                                                                  &                                                                                & \multicolumn{4}{c}{\textbf{Anomaly Detection Method (AUROC) }}                                                                                                                                                    &                                                                         \\ 
\cmidrule{3-6}
\begin{tabular}[c]{@{}l@{}}\textbf{Teacher}\\\textbf{Representation}\end{tabular} & \begin{tabular}[c]{@{}l@{}}\textbf{Classification}\\\textbf{Acc.}\end{tabular} & \begin{tabular}[c]{@{}l@{}}Knowledge\\Distillation\end{tabular} & MSE            & \begin{tabular}[c]{@{}l@{}}Mahalanobis \\(Diagonal)\end{tabular} & \begin{tabular}[c]{@{}l@{}}Mahalanobis\\(Full)\end{tabular} & \begin{tabular}[c]{@{}l@{}}\textbf{Avg L2}\\\textbf{Norm}\end{tabular}  \\ 
\midrule
\textit{Baseline}                                                                 & \textit{83.40}                                                                 & \textit{88.67}                                                  & \textit{88.51} & \textit{75.70}                                                   & \textit{85.11}                                              & \textit{0.126}                                                          \\ 
\midrule
Random                                                                            & 64.11                                                                          & 48.46                                                           & 52.50          & 52.03                                                            & 53.52                                                       & 0.064                                                                   \\ 
\midrule
STL Class.                                                                        & 71.28                                                                          & 58.99                                                           & 60.43          & 57.05                                                            & 55.59                                                       & 0.110                                                                   \\
CIFAR Class.                                                                      & 87.57                                                                          & 82.08                                                           & 90.74          & 77.34                                                            & 70.77                                                       & 0.142                                                                   \\ 
\midrule
STL RotNet                                                                        & 75.87                                                                          & 40.67                                                           & 41.11          & 43.82                                                            & 42.58                                                       & 0.063                                                                   \\
CIFAR RotNet                                                                      & 76.65                                                                          & 54.54                                                           & 51.11          & 51.26                                                            & 54.82                                                       & 0.077                                                                   \\
CvD RotNet                                                                        & 70.12                                                                          & 49.39                                                           & 50.71          & 50.35                                                            & 49.00                                                       & 0.062                                                                   \\ 
\midrule
STL AE                                                                            & 64.66                                                                          & 59.71                                                           & 55.98          & 54.62                                                            & 59.22                                                       & 0.023                                                                   \\
CIFAR AE                                                                          & 64.11                                                                          & 59.31                                                           & 54.75          & 54.05                                                            & 59.37                                                       & 0.033                                                                   \\
CvD AE                                                                            & 64.47                                                                          & 50.67                                                           & 51.25          & 50.71                                                            & 50.38                                                       & 0.052                                                                   \\ 
\midrule
STL DAE                                                                           & 64.69                                                                          & 58.19                                                           & 57.37          & 55.04                                                            & 62.04                                                       & 0.017                                                                   \\
CIFAR DAE                                                                         & 66.13                                                                          & 58.61                                                           & 53.92          & 52.89                                                            & 59.57                                                       & 0.020                                                                   \\
CvD DAE                                                                           & 57.23                                                                          & 50.37                                                           & 51.64          & 51.08                                                            & 51.84                                                       & 0.055                                                                   \\
\bottomrule
\end{tabular}
\end{table*}

\begin{table*}
\centering
\caption{Anomaly detection results for the dog class in the Cats vs. Dogs dataset.}
\label{fig:cvdresdog}
\footnotesize
\begin{tabular}{lllllll} 
\toprule
\multicolumn{1}{c}{}                                                              & \multicolumn{1}{c}{}                                                           & \multicolumn{4}{c}{\textbf{Anomaly Detection Method (AUROC) }}                                                                                                                                                   & \multicolumn{1}{c}{}                                                    \\ 
\cmidrule{3-6}
\begin{tabular}[c]{@{}l@{}}\textbf{Teacher}\\\textbf{Representation}\end{tabular} & \begin{tabular}[c]{@{}l@{}}\textbf{Classification}\\\textbf{Acc.}\end{tabular} & \begin{tabular}[c]{@{}l@{}}Knowledge\\Distillation\end{tabular} & MSE            & \begin{tabular}[c]{@{}l@{}}Mahalanobis\\(Diagonal)\end{tabular} & \begin{tabular}[c]{@{}l@{}}Mahalanobis\\(Full)\end{tabular} & \begin{tabular}[c]{@{}l@{}}\textbf{Avg L2}\\\textbf{Norm}\end{tabular}  \\ 
\midrule
\textit{Baseline}                                                                 & \textit{83.40}                                                                 & \textit{90.07}                                                  & \textit{89.57} & \textit{76.54}                                                  & \textit{86.74}                                              & \textit{0.152}                                                          \\ 
\midrule
Random                                                                            & 64.11                                                                          & 53.28                                                           & 47.99          & 48.63                                                           & 48.83                                                       & 0.060                                                                   \\ 
\midrule
STL Class.                                                                        & 71.28                                                                          & 64.66                                                           & 65.96          & 60.56                                                           & 57.09                                                       & 0.119                                                                   \\
CIFAR Class.                                                                      & 87.57                                                                          & 81.88                                                           & 93.84          & 79.12                                                           & 79.05                                                       & 0.161                                                                   \\ 
\midrule
STL RotNet                                                                        & 75.87                                                                          & 69.15                                                           & 66.30          & 60.73                                                           & 65.34                                                       & 0.066                                                                   \\
CIFAR RotNet                                                                      & 76.65                                                                          & 58.83                                                           & 51.59          & 51.62                                                           & 56.87                                                       & 0.081                                                                   \\
CvD RotNet                                                                        & 70.12                                                                          & 50.97                                                           & 47.44          & 48.23                                                           & 50.28                                                       & 0.054                                                                   \\ 
\midrule
STL AE                                                                            & 64.66                                                                          & 44.35                                                           & 48.33          & 48.58                                                           & 44.31                                                       & 0.023                                                                   \\
CIFAR AE                                                                          & 64.11                                                                          & 43.94                                                           & 49.38          & 48.85                                                           & 42.83                                                       & 0.032                                                                   \\
CvD AE                                                                            & 64.47                                                                          & 51.90                                                           & 48.61          & 49.14                                                           & 50.88                                                       & 0.055                                                                   \\ 
\midrule
STL DAE                                                                           & 64.69                                                                          & 49.10                                                           & 45.31          & 46.73                                                           & 40.99                                                       & 0.017                                                                   \\
CIFAR DAE                                                                         & 66.13                                                                          & 48.23                                                           & 49.35          & 49.37                                                           & 43.43                                                       & 0.023                                                                   \\
CvD DAE                                                                           & 57.23                                                                          & 51.37                                                           & 49.11          & 49.27                                                           & 49.48                                                       & 0.051                                                                   \\
\bottomrule
\end{tabular}
\end{table*}

\begin{table*}
\centering
\caption{Anomaly detection results for unimodal CIFAR-10 configuration, averaged over each class.}
\label{fig:cifarresuni}
\footnotesize
\begin{tabular}{llllll} 
\toprule
                                                                                  & \multicolumn{4}{c}{\textbf{Anomaly Detection Method (AUROC) }}                                                                                                                                                   &                                                                         \\ 
\cmidrule{2-5}
\begin{tabular}[c]{@{}l@{}}\textbf{Teacher}\\\textbf{Representation}\end{tabular} & \begin{tabular}[c]{@{}l@{}}Knowledge\\Distillation\end{tabular} & MSE            & \begin{tabular}[c]{@{}l@{}}Mahalanobis\\(Diagonal)\end{tabular} & \begin{tabular}[c]{@{}l@{}}Mahalanobis\\(Full)\end{tabular} & \begin{tabular}[c]{@{}l@{}}\textbf{Avg L2}\\\textbf{Norm}\end{tabular}  \\ 
\midrule
\textit{Baseline}                                                                 & \textit{95.44}                                                  & \textit{94.83} & \textit{94.99}                                                  & \textit{89.96}                                              & \textit{0.164}                                                          \\ 
\midrule
Random                                                                            & 60.09                                                           & 58.73          & 58.69                                                           & 57.25                                                       & 0.053                                                                   \\ 
\midrule
STL Classification                                                                & 85.29                                                           & 88.45          & 89.05                                                           & 82.66                                                       & 0.090                                                                   \\
FMNIST Classification                                                             & 57.60                                                           & 56.53          & 56.58                                                           & 56.60                                                       & 0.054                                                                   \\ 
\midrule
STL RotNet                                                                        & 70.71                                                           & 70.28          & 70.00                                                           & 66.32                                                       & 0.056                                                                   \\
FMNIST RotNet                                                                     & 57.90                                                           & 54.91          & 55.32                                                           & 52.81                                                       & 0.050                                                                   \\
CIFAR RotNet                                                                      & 85.35                                                           & 85.99          & 85.98                                                           & 84.88                                                       & 0.046                                                                   \\ 
\midrule
CIFAR SimCLR                                                                      & 54.18                                                           & 33.84          & 34.36                                                           & 50.85                                                       & 0.023                                                                   \\ 
\midrule
STL AE                                                                            & 62.85                                                           & 61.16          & 62.17                                                           & 61.03                                                       & 0.024                                                                   \\
FMNIST AE                                                                         & 63.50                                                           & 60.77          & 59.82                                                           & 59.24                                                       & 0.021                                                                   \\
CIFAR AE                                                                          & 61.03                                                           & 60.82          & 61.74                                                           & 59.33                                                       & 0.025                                                                   \\ 
\midrule
STL DAE                                                                           & 60.21                                                           & 62.48          & 63.00                                                           & 63.33                                                       & 0.019                                                                   \\
FMNIST DAE                                                                        & 60.19                                                           & 58.98          & 59.42                                                           & 57.47                                                       & 0.020                                                                   \\
CIFAR DAE                                                                         & 60.06                                                           & 55.53          & 57.41                                                           & 58.52                                                       & 0.022                                                                   \\
\bottomrule
\end{tabular}
\end{table*}

\begin{table*}
\centering
\caption{Anomaly detection results for multimodal CIFAR-10 configuration, averaged over each class.}
\label{fig:cifarresmulti}
\footnotesize
\begin{tabular}{llllll} 
\toprule
                                                                                  & \multicolumn{4}{c}{\textbf{Anomaly Detection Method (AUROC) }}                                                                                                                                                   &                                                                         \\ 
\cmidrule{2-5}
\begin{tabular}[c]{@{}l@{}}\textbf{Teacher}\\\textbf{Representation}\end{tabular} & \begin{tabular}[c]{@{}l@{}}Knowledge\\Distillation\end{tabular} & MSE            & \begin{tabular}[c]{@{}l@{}}Mahalanobis\\(Diagonal)\end{tabular} & \begin{tabular}[c]{@{}l@{}}Mahalanobis\\(Full)\end{tabular} & \begin{tabular}[c]{@{}l@{}}\textbf{Avg L2}\\\textbf{Norm}\end{tabular}  \\ 
\midrule
\textit{Baseline}                                                                 & \textit{87.60}                                                  & \textit{57.20} & \textit{57.39}                                                  & \textit{69.57}                                              & \textit{0.094}                                                          \\ 
\midrule
Random                                                                            & 50.98                                                           & 51.16          & 51.10                                                           & 51.13                                                       & 0.059                                                                   \\ 
\midrule
STL Class.                                                                        & 76.28                                                           & 54.03          & 55.81                                                           & 63.33                                                       & 0.128                                                                   \\
FMNIST Class.                                                                     & 51.08                                                           & 50.70          & 50.77                                                           & 50.81                                                       & 0.081                                                                   \\ 
\midrule
STL RotNet                                                                        & 55.31                                                           & 52.96          & 52.83                                                           & 52.90                                                       & 0.057                                                                   \\
FMNIST RotNet                                                                     & 51.46                                                           & 50.37          & 50.54                                                           & 50.52                                                       & 0.087                                                                   \\
CIFAR RotNet                                                                      & 61.35                                                           & 59.90          & 60.16                                                           & 53.59                                                       & 0.048                                                                   \\ 
\midrule
CIFAR SimCLR                                                                      & 49.37                                                           & 31.19          & 31.31                                                           & 43.45                                                       & 0.039                                                                   \\ 
\midrule
STL AE                                                                            & 51.47                                                           & 51.12          & 50.77                                                           & 51.54                                                       & 0.016                                                                   \\
FMNIST AE                                                                         & 51.89                                                           & 51.16          & 51.46                                                           & 51.29                                                       & 0.018                                                                   \\
CIFAR AE                                                                          & 51.64                                                           & 51.01          & 51.55                                                           & 51.22                                                       & 0.020                                                                   \\ 
\midrule
STL DAE                                                                           & 51.26                                                           & 51.36          & 51.61                                                           & 52.04                                                       & 0.009                                                                   \\
FMNIST DAE                                                                        & 52.31                                                           & 51.11          & 51.14                                                           & 51.06                                                       & 0.017                                                                   \\
CIFAR DAE                                                                         & 50.84                                                           & 50.36          & 50.04                                                           & 50.99                                                       & 0.013                                                                   \\
\bottomrule
\end{tabular}
\end{table*}

\begin{table*}
\centering
\caption{Complete Plant Pathology results.}
\label{fig:plantres2}
\footnotesize
\begin{tabular}{lllllll} 
\toprule
\multicolumn{1}{c}{}                                                              &                                                                                & \multicolumn{4}{c}{\textbf{Anomaly Detection Method (AUROC)}}                                                                                                                                                    &                                                                         \\ 
\cmidrule{3-6}
\begin{tabular}[c]{@{}l@{}}\textbf{Teacher}\\\textbf{Representation}\end{tabular} & \begin{tabular}[c]{@{}l@{}}\textbf{Classification}\\\textbf{Acc.}\end{tabular} & \begin{tabular}[c]{@{}l@{}}Knowledge\\Distillation\end{tabular} & MSE            & \begin{tabular}[c]{@{}l@{}}Mahalanobis\\(Diagonal)\end{tabular} & \begin{tabular}[c]{@{}l@{}}Mahalanobis\\(Full)\end{tabular} & \begin{tabular}[c]{@{}l@{}}\textbf{Avg L2}\\\textbf{Norm}\end{tabular}  \\ 
\midrule
\textit{Baseline }                                                                & \textit{100}                                                                   & \textit{100}                                                    & \textit{99.87} & \textit{99.89}                                                  & \textit{99.89}                                              & 0.00651                                                                 \\ 
\midrule
Random                                                                            & 63.67                                                                          & 42.50                                                           & 44.16          & 44.07                                                           & 41.35                                                       & 0.00114                                                                 \\ 
\midrule
Plant Village Class.                                                              & 90.92                                                                          & 89.82                                                           & 76.57          & 83.73                                                           & 90.78                                                       & 0.00436                                                                 \\
ImageNet Class.                                                                   & 88.24                                                                          & 70.66                                                           & 56.69          & 61.85                                                           & 49.45                                                       & 0.00356                                                                 \\ 
\midrule
Plant Path. RotNet                                                                & 57.97                                                                          & 48.09                                                           & 46.70          & 47.14                                                           & 47.51                                                       & 0.00097                                                                 \\
\bottomrule
\end{tabular}
\end{table*}

\begin{table*}
\centering
\caption{Complete X-ray results.}
\label{fig:xrayres2}
\footnotesize
\begin{tabular}{lllllll} 
\toprule
\multicolumn{1}{c}{}                                                              &                                                                                & \multicolumn{4}{c}{\textbf{Anomaly Detection Method (AUROC)}}                                                                                                                                                     &                                                                         \\ 
\cmidrule{3-6}
\begin{tabular}[c]{@{}l@{}}\textbf{Teacher}\\\textbf{Representation}\end{tabular} & \begin{tabular}[c]{@{}l@{}}\textbf{Classification}\\\textbf{Acc.}\end{tabular} & \begin{tabular}[c]{@{}l@{}}Knowledge\\Distillation\end{tabular} & MSE             & \begin{tabular}[c]{@{}l@{}}Mahalanobis\\(Diagonal)\end{tabular} & \begin{tabular}[c]{@{}l@{}}Mahalanobis\\(Full)\end{tabular} & \begin{tabular}[c]{@{}l@{}}\textbf{Avg L2}\\\textbf{Norm}\end{tabular}  \\ 
\midrule
\textit{Baseline }                                                                & \textit{98.98 }                                                                & \textit{99.74}                                                  & \textit{99.25 } & \textit{99.68 }                                                 & \textit{99.63 }                                             & 0.00418                                                                 \\ 
\midrule
Random                                                                            & 88.56                                                                          & 73.61                                                           & 42.41           & 44.85                                                           & 71.87                                                       & 0.00120                                                                 \\ 
\midrule
ImageNet Class.                                                                   & 96.79                                                                          & 76.36                                                           & 68.37           & 69.35                                                           & 73.46                                                       & 0.00383                                                                 \\ 
\midrule
X-ray RotNet                                                                      & 93.62                                                                          & 96.41                                                           & 36.08           & 37.77                                                           & 60.06                                                       & 0.00453                                                                 \\
\bottomrule
\end{tabular}
\end{table*}

\end{appendices}
\end{document}